\documentclass{article}

\usepackage{arxiv}
\usepackage{graphicx}
\graphicspath{ {./images/} }

\usepackage{graphics, pdflscape, caption, xcolor}
\usepackage{hyperref, url, cite}
\usepackage{amsmath,amssymb,amsfonts}
\usepackage{algorithmic, textcomp, footnote, booktabs}
\usepackage{orcidlink}

\fancypagestyle{fancyfirst}{
    \fancyhf{}

    \fancyfoot[C]{
        \parbox{\textwidth}{
            \centering
            \scriptsize
            \textbf{Accepted for publication in:}
            \textit{Latin American Conference on Informatics (CLEI 2026)}\\[2pt]

            \textbf{Research Repository:}
            \href{https://github.com/jvalverr/sentimex-llm-benchmark/tree/main/cultural_reasoning}
            {SentiMEX-LLM Benchmark -- Cultural Reasoning}\\[2pt]

            \textbf{Research Project:}
            \href{https://jvalverr.github.io/projects/03_digital_media_observatory/}
            {Digital Media Observatory}\\[4pt]

            \thepage
        }
    }
}

\title{Evaluating Cultural Reasoning and Robustness in LLMs for Spanish Political Discourse}

\author{
 Jorge Valverde-Rebaza\thanks{Corresponding author: jvalverr@tec.mx}\orcidlink{0000-0001-8664-9692} \\
  School of Engineering and Sciences \\
  Tecnologico de Monterrey\\
  Mexico \\
  \texttt{jvalverr@tec.mx} \\
  \And
 Nathan Garcia-Larios\orcidlink{0009-0007-0469-1514} \\
  School of Engineering and Sciences \\
  Tecnologico de Monterrey\\
  Mexico \\
  \texttt{A01749595@tec.mx} \\
\And
   Karime Ruvalcaba-Pérez\orcidlink{0009-0007-9372-0435}\\
  School of Engineering and Sciences \\
  Tecnologico de Monterrey\\
  Mexico \\
  \texttt{A01656188@tec.mx} \\
\And
 Alejandro Rendón\orcidlink{0009-0000-6531-4471}\\
  School of Engineering and Sciences \\
  Tecnologico de Monterrey\\
  Mexico \\
  \texttt{A01664589@tec.mx} \\
\And
 Santiago Sánchez-Ballesteros\orcidlink{0009-0006-1437-1509}\\
  School of Engineering and Sciences \\
  Tecnologico de Monterrey\\
  Mexico \\
  \texttt{A01656188@tec.mx} \\
\And
 Alejandro Martin del Campo\orcidlink{0000-0001-9584-2486}\\
  School of Engineering and Sciences \\
  Tecnologico de Monterrey\\
  Mexico \\
  \texttt{martindelcampo@tec.mx} \\
}

\begin{document}
\maketitle
\thispagestyle{fancyfirst}

\begin{abstract}
Large language models (LLMs) are increasingly used to interpret, generate, and mediate public-facing language, which makes their reliability especially consequential in politically sensitive settings. Although sentiment analysis is a common task in this area, affective meaning in social media discourse is often conveyed through culturally situated and pragmatically indirect cues that are not well captured by label-centric evaluation alone. In this paper, we evaluate the robustness of LLMs for sentiment analysis of Spanish political discourse from social media, with emphasis on culturally situated affective interpretation. We introduce a diagnostic protocol that combines three independent human annotation views into Gold and Ambiguous subsets, and extends the corpus with a Synthetic subset built from sentiment-flipped counterfactuals selected through model-conditioned perplexity and LLM-as-a-judge criteria. Using this framework, we compare proprietary and open-weight models under zero-shot and few-shot prompting, focusing on predictive robustness across human agreement conditions and counterfactual shift. Our results show that annotation ambiguity and counterfactual perturbation reveal failure modes that conventional evaluation on the original corpus tends to obscure. GPT-4.1-mini remains the most stable among the evaluated models, whereas the two smaller open-weight models degrade more markedly under synthetic perturbations, and few-shot prompting provides only limited gains. Overall, the paper contributes a reproducible protocol for robustness-oriented evaluation in political sentiment analysis, an empirical characterization of model behavior under ambiguity and sentiment-flipped counterfactuals, and a publicly available Spanish social media dataset with three-way human sentiment annotations.

\end{abstract}


\section{Introduction}
\label{sec:intro}

Large language models (LLMs) have rapidly become foundational components of contemporary language technologies, supporting tasks such as classification, summarization, explanation generation, content moderation, retrieval support, and synthetic text production~\cite{LLMs:systematic:survey:24}. As their deployment expands across education, industry, and public-sector workflows, so does their capacity to shape how information is produced, interpreted, and circulated. This is particularly consequential in democratic contexts, where language is not merely a medium of communication but a core mechanism of participation, deliberation, and accountability~\cite{jungherr2023ai}. In this sense, LLMs may strengthen the public sphere by improving access to information, facilitating citizen-government interaction, supporting civic engagement, and helping identify citizen-reported issues from large-scale social media data~\cite{Lazar:LLMs:democracy:26,Santos:citizen:LLMs:24}. At the same time, their generative and persuasive capabilities create new risks, including the large-scale production of convincing political content, the simulation of citizen-like discourse, and the amplification of misinformation or disinformation narratives that may affect public trust and democratic representation~\cite{hackenburg2025levers,Qazi:LLMs:factchecking:26,vykopal:LLM:disinformation:24}.

Affective signals are central to this problem because political communication often seeks not only to inform, but also to mobilize, persuade, polarize, or delegitimize actors and institutions. Within natural language processing (NLP), sentiment analysis is a long-standing task for identifying opinions, affective orientation, and evaluative stance in text, and it has been widely used to summarize large volumes of user-generated content~\cite{Chandan:NLP:survey:2025}. In political and participatory settings, these methods can support the analysis of online discussions, public concerns, and deliberative processes~\cite{behrendt:ML:political:discussion:26}. Prior work has also advanced sentiment analysis for Spanish social media in electoral contexts, showing the relevance of annotated corpora and LLM-based evaluation for politically charged discourse~\cite{BRITO:latinamerica_elections:23,Valverde:LLMs_sentiment:classif:26}. However, political sentiment in social media is rarely reducible to stable labels or surface polarity. Its interpretation often depends on tone, irony, emojis, hashtags, local references, and culturally shared pragmatic cues. Thus, strong performance under standard metrics may conceal deeper interpretive limitations, especially when models face ambiguous or culturally situated language. This aligns with recent arguments that evaluating generative AI should be treated as a measurement challenge, where the target construct and the validity of the evaluation instrument must be explicitly examined~\cite{Wallach:eval:GenAI:social_sciece:25,ochieng2026reasoning}.

Although recent work has produced increasingly sophisticated protocols for improving and evaluating formal multi-step reasoning in LLMs, including mathematical reasoning through prompting, chain-of-thought variants, and self-reflection mechanisms~\cite{Wang:LLM_math_reasoning_survey26,Loureiro:LLMs_reasoning_maps:26}, analogous protocols for culturally situated reasoning remain less mature~\cite{Yu:Language_reasoning:24}. Emerging research on cultural awareness, cultural alignment, and cultural benchmarks has shown that multilingual competence does not imply cultural competence, and that LLMs remain brittle when reasoning about culture-specific knowledge, social norms, and situated meanings~\cite{alkhamissi2024cultural,chiu2025culturalbench,Pawar:LLMs_cultural_awareness:25}.

Evaluation practice itself is also changing. The LLM-as-a-judge paradigm has emerged as a scalable alternative for scoring, ranking, and selecting open-ended outputs, and has been shown to approximate human preferences reasonably well in some settings~\cite{zheng2023judging}. Yet judge models are not neutral evaluators: they can exhibit systematic biases, remain vulnerable to perturbations, and require careful reliability analysis and protocol design~\cite{chen2024judgebias,li2025generation}. For this reason, LLM-based judgment is best treated not as a replacement for human grounding, but as one component within a broader evaluation framework. This motivates evaluation designs that combine human grounding with model-assisted assessment rather than relying exclusively on either source.

In this paper, we build on these lines of work to study whether LLMs maintain culturally situated affective inferences when Spanish political language is humanly ambiguous or perturbed through counterfactual rewrites. We focus on Spanish social media discourse and move beyond conventional label-centric evaluation by combining three independent human annotation views with perturbation-based stress testing. Concretely, we construct a \emph{Gold} subset from full annotator agreement, an \emph{Ambiguous} subset from partial or complete annotator disagreement, and a \emph{Synthetic} subset composed of sentiment-flipped counterfactuals generated from non-neutral posts. Synthetic instances retain the agreement condition of their source posts, allowing us to compare model behavior on original Gold and Ambiguous instances with counterfactual variants derived from them. We then compare GPT-4.1-mini, Llama3.1:8b, and DeepSeek-r1:8b under zero-shot and few-shot prompting, focusing on predictive robustness across human agreement conditions and counterfactual shift.

Our contributions are threefold:

\begin{itemize}
    \item We adapt and operationalize an evaluation protocol for Spanish political sentiment analysis that uses multi-annotator consensus, annotation ambiguity, and counterfactual perturbation as diagnostic conditions for assessing LLM robustness.
    \item We empirically characterize robustness differences between proprietary and open-weight LLMs under zero-shot and few-shot prompting, showing how performance changes across Gold, Ambiguous, and Synthetic conditions.
    \item We will release the human sentiment annotations, evaluation artifacts, and Spanish political discourse dataset.
\end{itemize}

The remainder of this paper is organized as follows. Section~\ref{sec:related} reviews related work on cultural reasoning, political sentiment analysis, and model-assisted evaluation. Section~\ref{sec:eval_protocol} presents the evaluation protocol, including the dataset, agreement-based subset construction, and counterfactual generation and selection. Section~\ref{sec:experiment} describes the evaluated models, prompting settings, and metrics. Section~\ref{sec:results} reports and discusses the main findings regarding annotation ambiguity, counterfactual shift, and prompting strategy. Finally, Section~\ref{sec:conclusion} concludes the paper and discusses implications for robust and culturally aware LLM evaluation.
Show all of the project history or only labelled versions.

\section{Related Work}
\label{sec:related}

This work is grounded in two complementary research lines. The first concerns reasoning-aware evaluation and model-assisted assessment of LLM outputs. Recent work has shown that evaluating LLMs is increasingly difficult because their outputs are open-ended, context-dependent, and not always well captured by single-task accuracy measures~\cite{LLMs:systematic:survey:24}. More broadly, natural language reasoning involves heterogeneous forms of inference over unstructured text, which makes evaluation especially challenging when tasks require more than surface-level pattern recognition~\cite{Yu:Language_reasoning:24}. In this context, the LLM-as-a-judge paradigm has emerged as a scalable mechanism for scoring, ranking, and selecting generated outputs. Zheng et al.~\cite{zheng2023judging} showed that strong LLM judges can approximate human preferences in conversational evaluation, making model-based assessment useful when human evaluation is costly. However, subsequent studies have emphasized that judge models are not neutral instruments. Chen et al.~\cite{chen2024judgebias} show that both human and LLM judges may exhibit systematic biases, while Li et al.~\cite{li2025generation} discuss broader opportunities and limitations of using LLMs as evaluators. These findings motivate protocols in which model-assisted judgment may support candidate selection and diagnostic evaluation, but must remain constrained, documented, and grounded in human-annotated data.

The second line concerns cultural reasoning, cultural alignment, and politically sensitive language. Work on cultural alignment has shown that LLM behavior varies with prompting language, training mixtures, and cultural context~\cite{alkhamissi2024cultural}. Chiu et al.~\cite{chiu2025culturalbench} further show that even strong models remain limited on diverse, human-authored cultural knowledge tasks, reinforcing Pawar et al.'s~\cite{Pawar:LLMs_cultural_awareness:25} view that cultural awareness requires more than multilingual capability, including sensitivity to social norms, values, and situated context. This gap is especially relevant because recent progress in LLM reasoning has focused heavily on mathematical or formal multi-step reasoning mechanisms~\cite{Wang:LLM_math_reasoning_survey26,Loureiro:LLMs_reasoning_maps:26}. By contrast, culturally situated reasoning remains less mature as an evaluation target, particularly when the task involves affective interpretation rather than factual cultural knowledge.

Sentiment analysis in political discourse provides a concrete setting where these issues converge. As a well-established task for identifying opinions and affective orientation in text~\cite{Chandan:NLP:survey:2025}, it has been widely applied to social media data for analyzing political behavior and electoral dynamics in Latin America~\cite{BRITO:latinamerica_elections:23}. In Spanish electoral social media, prior work has already introduced annotated resources and benchmarked pre-trained and large language models mainly through predictive sentiment performance~\cite{Valverde:LLMs_sentiment:classif:26}. The present paper takes this line of research in a different direction by shifting the emphasis from standard label prediction to robustness under annotator ambiguity and counterfactual perturbation. Methodologically, it builds on Ochieng et al.~\cite{ochieng2026reasoning}, who argue that sentiment in culturally nuanced contexts should be evaluated beyond labels through ambiguity and counterfactual analysis. This use of counterfactual perturbation also connects to broader work on counterfactual reasoning and generation, where minimally altered rewrites are used to examine whether models preserve relevant invariances under controlled changes~\cite{qin:counterfactual:19}. By adapting this perspective to Spanish political social media discourse, we position our work at the intersection of model-assisted evaluation and cultural reasoning, examining how ambiguity and counterfactual shifts affect predictive robustness in culturally situated sentiment classification.

\section{Evaluation Protocol}
\label{sec:eval_protocol}

This section describes the protocol used to transform a Spanish political sentiment corpus into a diagnostic evaluation setting for LLM robustness and culturally situated affective reasoning. The protocol treats human consensus, human ambiguity, and counterfactual shift as complementary evaluation conditions for assessing whether models go beyond label prediction and remain consistent under culturally and pragmatically sensitive variation.


\subsection{Dataset and Human Annotation}
\label{sec:eval_protocol:data_annot}

We use a Spanish political social media corpus collected between January 1 and July 31, 2024, from Facebook, Instagram, X, and YouTube, originally introduced by Valverde-Rebaza et al.~\cite{Valverde:LLMs_sentiment:classif:26}. The corpus contains posts related to the 2024 Mexican electoral process and was annotated independently by three native Spanish-speaking annotators using three sentiment categories: positive, negative, and neutral. In this paper, we use the resulting three-way annotation matrix as the basis for diagnostic evaluation, while the details of data collection, filtering, annotation guidelines, and inter-annotator agreement are provided in the original corpus description.

The resulting dataset comprises $1,354$ annotated posts distributed across the four platforms, as summarized in Table~\ref{tab:data_distribution}. In the table, POS, NEG, and NEU denote the consolidated sentiment labels; GOLD and AMBIGUOUS refer to the agreement-based subsets derived from the three human annotations; and SYNTHETIC denotes the non-neutral instances used for counterfactual generation.

\begin{table*}[!t]
\centering
\caption{Distribution of sentiment labels and evaluation subsets by platform.}
\label{tab:data_distribution}

\setlength{\tabcolsep}{12pt}
\renewcommand{\arraystretch}{1.25}

\begin{tabular}{lrrrrrrr}
\toprule
\textbf{Platform} &
\textbf{POS} &
\textbf{NEG} &
\textbf{NEU} &
\textbf{GOLD} &
\textbf{AMBIGUOUS} &
\textbf{SYNTHETIC} &
\textbf{TOTAL} \\
\midrule
Facebook  & 144 &  99 &  95 & 234 & 104 & 243 & 338 \\
Instagram & 193 &  52 &  97 & 220 & 122 & 245 & 342 \\
X         &  72 & 276 &  31 & 253 & 126 & 348 & 379 \\
YouTube   &  14 &  67 & 214 & 142 & 153 &  81 & 295 \\
\midrule
\textbf{Total} &
\textbf{423} &
\textbf{494} &
\textbf{437} &
\textbf{849} &
\textbf{505} &
\textbf{917} &
\textbf{1,354} \\
\bottomrule
\end{tabular}

\end{table*}

When considering platform-level composition, X provides the largest subset, with $379$ posts, followed by Instagram with $342$, Facebook with $338$, and YouTube with $295$. The sentiment composition varies substantially across platforms. Facebook is comparatively balanced, with $144$ positive, $99$ negative, and $95$ neutral posts, whereas Instagram is skewed toward positive sentiment, with $193$ positive posts against $52$ negative and $97$ neutral instances. X exhibits a sharply polarized profile toward negative sentiment, with $276$ negative posts and a much smaller presence of $72$ positive and $31$ neutral instances. YouTube, in turn, is dominated by neutral content, with $214$ neutral posts, $67$ negative posts, and only $14$ positive posts. These differences indicate that the corpus does not represent a homogeneous affective space across platforms. Accordingly, while platform-level variation is reported for transparency, the evaluation in this paper uses agreement structure and counterfactual perturbation as the primary diagnostic axes.

\begin{figure}[h]
    \centering
    \includegraphics[width=\linewidth]{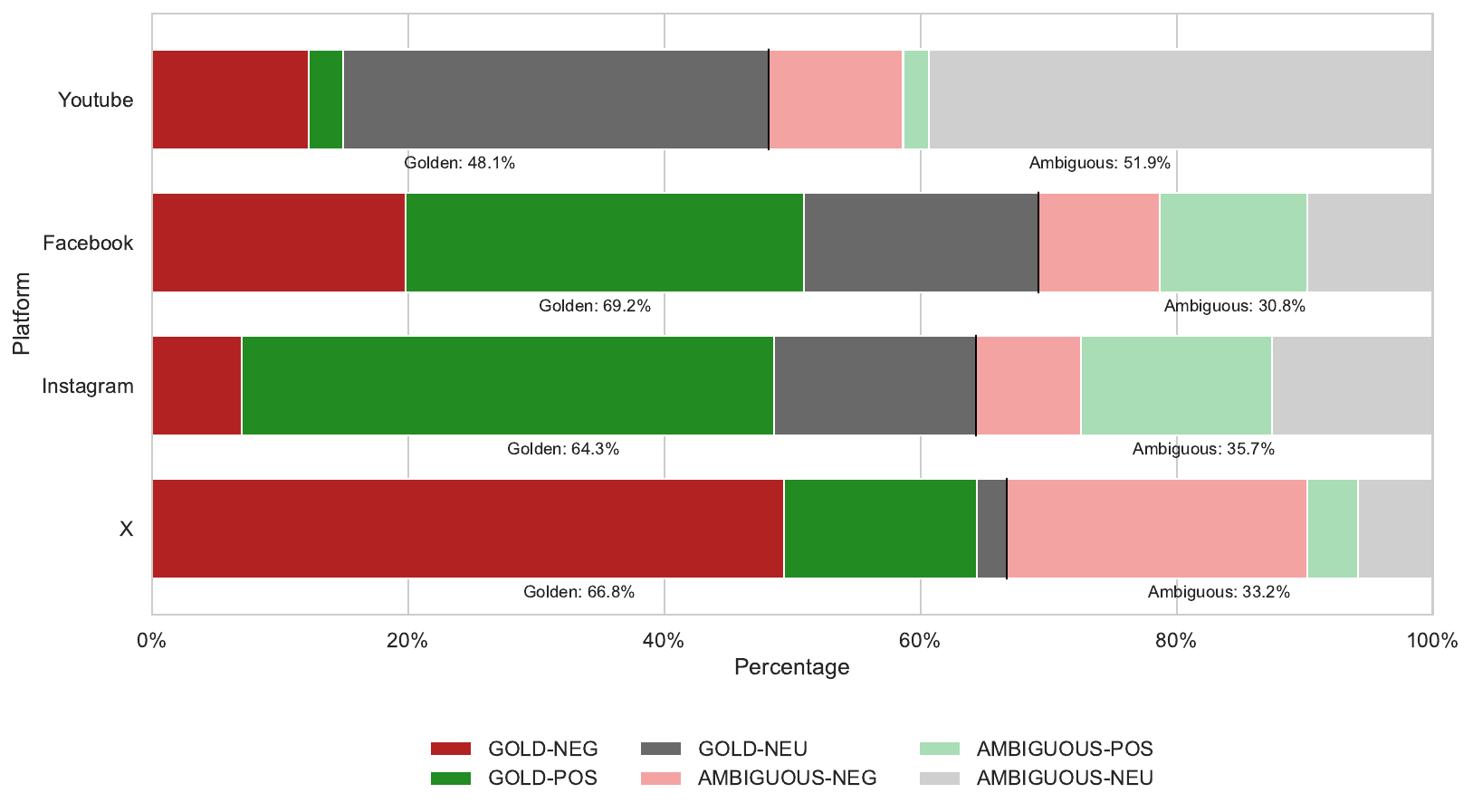}
    \caption{Distribution of Gold and Ambiguous instances by platform and sentiment label. The Synthetic subset is constructed from non-neutral Gold and Ambiguous instances through sentiment-flipped counterfactuals.}
    \label{fig:platform_agreement_sentiment}
\end{figure}

Figure~\ref{fig:platform_agreement_sentiment} complements Table~\ref{tab:data_distribution} by visualizing the platform-level distribution of Gold and Ambiguous instances across sentiment categories. The figure shows that ambiguity is not uniformly distributed, with YouTube showing the highest proportion of ambiguous cases, whereas Facebook, Instagram, and X concentrate a larger share of Gold instances. This supports our use of agreement structure as a central diagnostic axis in the evaluation protocol.

\subsection{Gold and Ambiguous Subsets}
\label{sec:eval_protocol:gold}

The three independent annotations associated with each post allow us to derive two agreement-based subsets, reported in Table~\ref{tab:data_distribution}. The \emph{Gold} subset includes posts for which the three annotators assigned the same sentiment label, representing cases of full human consensus. The \emph{Ambiguous} subset includes all posts without full agreement, covering both partial agreement cases, in which two annotators selected the same label and one disagreed, and complete disagreement cases, in which the three annotators selected different labels.

Overall, the Gold subset contains $849$ posts, corresponding to $62.7\%$ of the annotated corpus, while the Ambiguous subset contains $505$ posts, corresponding to $37.3\%$. At the platform level, Facebook includes $234$ Gold and $104$ Ambiguous posts, Instagram includes $220$ Gold and $122$ Ambiguous posts, X includes $253$ Gold and $126$ Ambiguous posts, and YouTube includes $142$ Gold and $153$ Ambiguous posts. As shown in Figure~\ref{fig:platform_agreement_sentiment}, YouTube is the only platform where Ambiguous instances slightly exceed Gold instances, whereas Facebook, Instagram, and X show larger proportions of Gold cases.

This partition is central to the evaluation protocol. Gold instances serve as the reference condition for cases in which sentiment interpretation is stable across human annotators. Ambiguous instances, in contrast, are treated as a diagnostic condition rather than as annotation noise. In political social media discourse, disagreement may reflect the presence of implicit stance, irony, platform-specific conventions, or context-dependent affective cues. For this reason, comparing model behavior across Gold and Ambiguous subsets allows us to evaluate whether LLMs remain robust when the sentiment signal is less explicit and less consistently interpreted by humans.


\subsection{Counterfactual Generation and Selection}
\label{sec:eval_protocol:counterfactual}

\begin{landscape}
\begin{figure}[p]
    \centering
    \includegraphics[
        width=0.98\linewidth,
        height=0.88\textheight,
        keepaspectratio,
        trim={0 5.8cm 0 1.8cm},
        clip
    ]{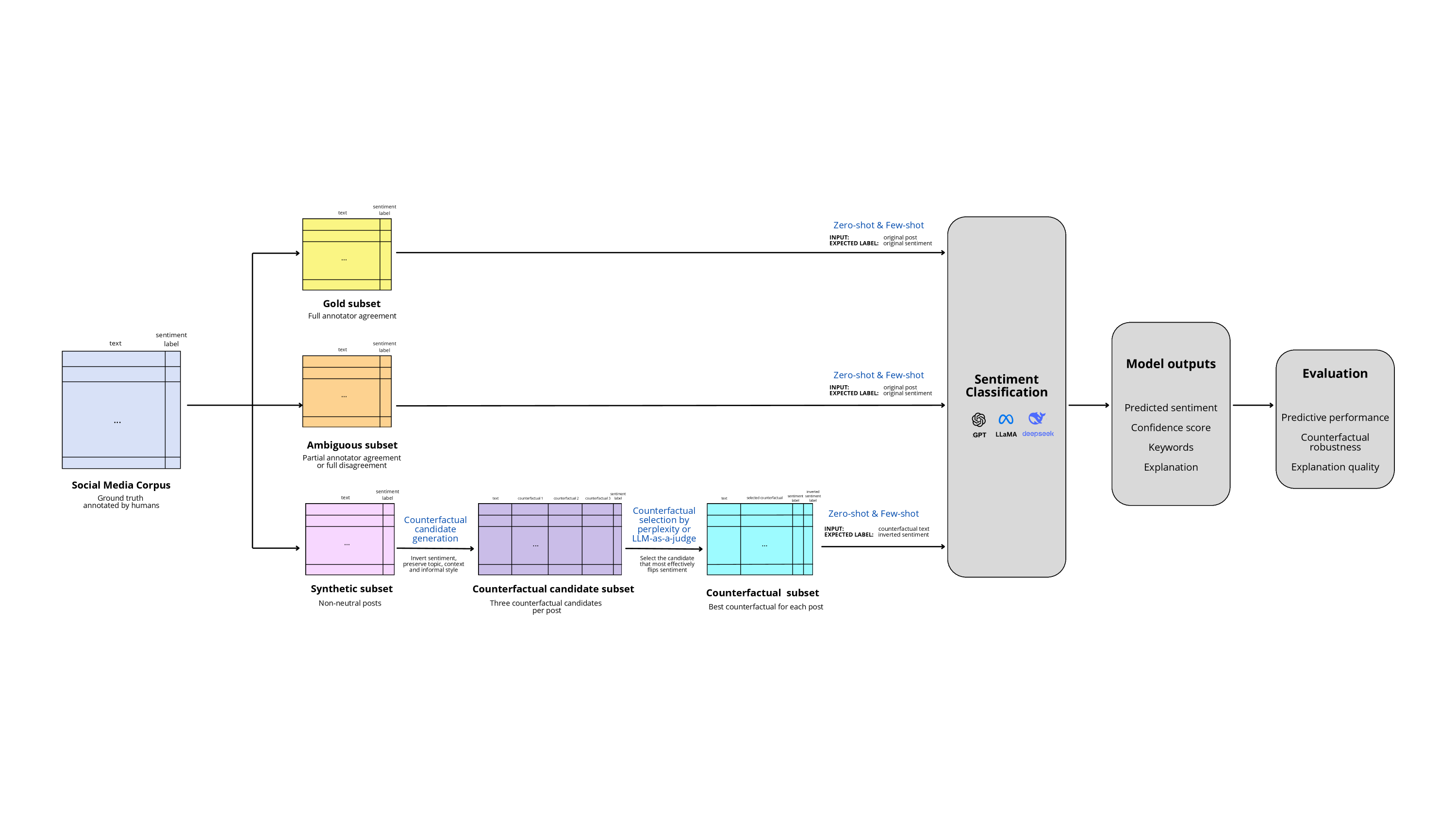}

    \caption{Experimental flow for evaluating cultural reasoning in political sentiment analysis. Gold and Ambiguous instances are classified using the original posts and evaluated against the original sentiment label, whereas Synthetic instances are first constructed through counterfactual generation and selection and then evaluated against the expected inverted sentiment.}
    \label{fig:process_cultural_reasoning}
\end{figure}
\end{landscape}

To evaluate robustness under controlled affective perturbation, we construct a \emph{Synthetic} subset composed of sentiment-flipped counterfactuals. In line with reasoning-aware evaluation in culturally nuanced sentiment analysis~\cite{ochieng2026reasoning} and broader work on counterfactual rewriting~\cite{qin:counterfactual:19}, these counterfactuals are used to test whether models respond consistently when the affective orientation of a post is intentionally modified while its topic, informal style, and political context are preserved. Figure~\ref{fig:process_cultural_reasoning} summarizes the complete evaluation flow, distinguishing the direct classification path followed by Gold and Ambiguous instances from the counterfactual construction path followed by Synthetic instances.

Counterfactual generation is applied only to non-neutral posts, since polarity inversion is directly defined for positive and negative instances but less straightforward for neutral content. As reported in Table~\ref{tab:data_distribution}, the Synthetic subset is derived from $917$ eligible source instances, corresponding to the $423$ positive and $494$ negative posts in the corpus. At the platform level, X contributes the largest number of eligible source posts, with $348$ non-neutral instances, followed by Instagram with $245$, Facebook with $243$, and YouTube with $81$. This distribution reflects the sentiment profile of each platform, since X contributes many counterfactual sources due to its strong concentration of negative posts, whereas YouTube contributes fewer because most of its content is neutral.

For each eligible post, the counterfactual construction process is model-conditioned. That is, for each model-specific run, the same LLM associated with that run is prompted to generate three alternative rewrites that invert the original sentiment. Although Figure~\ref{fig:process_cultural_reasoning} abstracts this step as counterfactual candidate generation, the generation itself is performed by the corresponding LLM under an explicit rewriting prompt. The transformation may involve changes in sentiment-bearing keywords, negation, tone, framing, emojis, hashtags, or sentence structure. The objective is not to generate unrestricted paraphrases, but to create plausible variants that alter the affective orientation while preserving the communicative setting of the original post. This model-conditioned design means that the Synthetic condition should be interpreted as an end-to-end diagnostic of each model's ability to generate, select, and classify sentiment-flipped variants, rather than as an externally validated counterfactual benchmark.

The resulting candidates are then filtered through two complementary selection strategies. The first is based on perplexity and selects the candidate with the lowest model-based expectedness score, using perplexity as a proxy for linguistic plausibility within the corresponding model run. The second uses an LLM-as-a-judge criterion, where the same model-conditioned LLM selects the candidate that best preserves fluency, plausibility, contextual consistency, and clarity of the sentiment shift. Thus, although the figure presents this stage compactly as counterfactual selection, the judge-based selection stage is also model-assisted. These criteria are kept separate because they capture different aspects of counterfactual suitability: perplexity favors linguistically expected formulations, whereas the judge-based criterion prioritizes communicative adequacy and the explicitness of the affective inversion.

\clearpage
\clearpage

After selection, the chosen counterfactual becomes the input for the Synthetic classification condition. In this condition, the model does not classify the original post; instead, it classifies the selected counterfactual text, and the expected label is the inverted sentiment relative to the original non-neutral post. For example, a positive original post is expected to become negative after counterfactual rewriting, and a negative original post is expected to become positive. Synthetic instances also retain the agreement condition of their source posts. Thus, Synthetic-Gold refers to counterfactuals generated from non-neutral posts originally belonging to the Gold subset, whereas Synthetic-Ambiguous refers to counterfactuals generated from non-neutral posts originally belonging to the Ambiguous subset. This distinction allows us to compare model behavior on original posts and counterfactual variants while preserving the role of human agreement as a diagnostic axis.

\section{Experimental Setup}
\label{sec:experiment}

This section describes how the proposed protocol was instantiated across models, prompting strategies, and evaluation metrics. The experiments compare model behavior over the Gold, Ambiguous, and Synthetic conditions, focusing on sentiment classification robustness under human agreement, annotator ambiguity, and counterfactual shift. In all conditions and as shown in Figure~\ref{fig:process_cultural_reasoning}, the target task remains sentiment prediction; what changes is the input type and the reference label used for evaluation. Gold and Ambiguous instances are classified using the original post and the consolidated sentiment label, whereas Synthetic instances are classified using the selected counterfactual text and the expected inverted sentiment label.

\subsection{Models}
\label{sec:experiment:models}

We evaluate three LLMs selected to cover proprietary and open-weight deployment settings. \texttt{GPT-4.1-mini} is used as the proprietary reference model. \texttt{Llama3.1:8b} is included as an open-weight instruction-tuned model, providing a reproducible baseline for local deployment. \texttt{DeepSeek-r1:8b} is included as an open-weight reasoning-oriented model, allowing us to examine whether reasoning-specialized behavior is reflected in sentiment interpretation under ambiguity and counterfactual shift.

The comparison should be interpreted with attention to model scale and availability. \texttt{Llama3.1:8b} and \texttt{DeepSeek-r1:8b} are 8B-parameter models and therefore correspond to relatively small LLMs. By contrast, \texttt{GPT-4.1-mini} is a proprietary model whose exact parameter count is not publicly documented. Consequently, the experiments should not be read as a general comparison between proprietary and open-weight models, nor as representative of larger frontier models or larger open-weight alternatives. Rather, they compare three accessible models under the same robustness-oriented diagnostic protocol.

All models were evaluated under the same diagnostic protocol and label space. Sentiment labels were encoded as three mutually exclusive classes: \emph{negative}, \emph{neutral}, and \emph{positive}. To reduce stochastic variation in classification and rubric-based scoring, temperature was set to $0.0$ for sentiment prediction and explanation evaluation. For counterfactual generation, temperature was increased to $0.9$ to encourage lexical and pragmatic diversity among candidate rewrites. Counterfactual selection was performed with a low-temperature setting to favor stable candidate choice.


\subsection{Prompting Settings}
\label{sec:experiment:prompting_setting}

We evaluate each model under zero-shot and few-shot prompting. In the zero-shot setting, the model receives the target post and the sentiment label definitions, without labeled examples. In the few-shot setting, the same classification instruction is augmented with labeled examples sampled from the annotated corpus. We use up to $k=3$ examples per class with a fixed random seed, yielding a balanced demonstration set whenever sufficient examples are available. The same sampling procedure is applied across models to preserve comparability.

All classification prompts require a structured JSON response containing the predicted sentiment label, a confidence score from $0$ to $5$, supporting keywords, and a brief explanation. In this paper, explanations are retained as part of the model output for traceability, but the quantitative analysis focuses on predictive robustness. This structured response format constrains the output to the predefined label space and reduces invalid or verbose generations.

For the Synthetic condition, prompting is used in two additional stages already defined in Section~\ref{sec:eval_protocol:counterfactual}. First, the counterfactual generation prompt asks the model to produce three sentiment-flipped rewrites of each eligible non-neutral post while preserving topic, political context, informal tone, and plausibility. Second, the judge-based selection prompt asks the corresponding model-conditioned LLM to choose the candidate that most clearly and plausibly expresses the intended sentiment inversion. The perplexity-based selection condition does not require an additional prompt, since it ranks candidates through model-based expectedness rather than judgment instructions.

Table~\ref{tab:prompt_templates} summarizes the core instructions implemented in the prompt templates used in the experimental pipeline. The table reports functionally equivalent descriptions rather than verbatim prompts, while the complete raw templates will be released with the experimental artifacts.

\begin{table*}[!t]
\centering
\caption{Functional summary of the prompt templates used in the experimental pipeline.}
\label{tab:prompt_templates}

\setlength{\tabcolsep}{10pt}
\renewcommand{\arraystretch}{1.35}

\begin{tabular}{p{0.23\textwidth}p{0.70\textwidth}}
\toprule
\textbf{Prompt} & \textbf{Core instruction and expected output} \\
\midrule

\textbf{Sentiment classification} &
Classify the input post as \emph{negative}, \emph{neutral}, or \emph{positive}. Return a JSON object containing the predicted label, supporting keywords, a brief explanation, and a confidence score from $0$ to $5$. \\

\addlinespace[5pt]

\textbf{Few-shot classification} &
Apply the same classification schema using labeled examples sampled from the annotated corpus. Include up to $k=3$ examples per sentiment class when available. \\

\addlinespace[5pt]

\textbf{Counterfactual generation} &
Generate three alternative rewrites of a non-neutral post that invert its sentiment while preserving language, topic, political context, informal tone, and plausibility. Return each candidate as a JSON object. \\

\addlinespace[5pt]

\textbf{Counterfactual selection} &
Select the candidate that most effectively inverts sentiment while preserving fluency, realism, and contextual consistency. Return the selected counterfactual, its predicted sentiment, and a brief justification. \\

\bottomrule
\end{tabular}

\end{table*}


\subsection{Metrics}
\label{sec:experiment:metrics}

The evaluation metrics are aligned with the three diagnostic conditions defined in the protocol. For Gold and Ambiguous instances, the model prediction is compared against the consolidated sentiment label of the original post. For Synthetic instances, the model prediction is compared against the expected inverted label of the selected counterfactual. Thus, the classification task remains sentiment prediction in all cases, but the Synthetic condition evaluates whether the model recognizes the intended polarity shift introduced by the counterfactual rewrite.

For sentiment prediction, we report F1-scores by class and evaluation condition. Results are computed separately for Gold, Ambiguous, and Synthetic subsets, allowing us to compare performance under full human agreement, annotator disagreement, and controlled sentiment shift. This separation is central to the experimental design because aggregate performance over the original corpus may obscure failures that emerge only when sentiment is ambiguous or deliberately perturbed.

For the Synthetic condition, we also use two auxiliary diagnostics to interpret the counterfactual construction process. The \emph{flip rate} measures whether the selected counterfactual changes the sentiment label relative to the original post. In this setting, a successful flip indicates that the rewritten text expresses the intended opposite polarity. We also compute token-level Levenshtein distance between the original post and the selected counterfactual, which estimates the magnitude of textual modification introduced by the rewrite. These measures are used as construction-level diagnostics, whereas the main comparative analysis is based on sentiment classification performance across models, prompting strategies, and evaluation conditions.

\section{Results and Discussion}
\label{sec:results}

This section reports the empirical results obtained under the proposed diagnostic protocol. We focus on three complementary views. First, we compare F1-scores across Gold, Ambiguous, and Synthetic conditions. Second, we contrast these results with a contextual reference derived from conventional original-corpus sentiment classification. Third, we analyze the drop in F1-score relative to the Original-Gold condition, which represents the clearest setting in terms of human agreement.

\subsection{F1 Performance Under Diagnostic Conditions}
\label{sec:results:f1}

Table~\ref{tab:f1_main_results} reports the F1-scores obtained by each model under zero-shot and few-shot prompting. The table separates the original classification setting from the two Synthetic conditions, in which counterfactuals were selected either by perplexity or by an LLM-as-a-judge criterion. Gold and Ambiguous refer to the agreement condition of the source post. Thus, in the Synthetic conditions, Synthetic-Gold corresponds to counterfactuals derived from non-neutral posts originally belonging to the Gold subset, whereas Synthetic-Ambiguous corresponds to counterfactuals derived from non-neutral posts originally belonging to the Ambiguous subset.

\begin{table*}[t]
\centering
\caption{F1-scores by model, prompting strategy, and diagnostic condition. Gold and Amb. refer to the agreement condition of the source post. Synthetic-PPL and Synthetic-Judge correspond to model-conditioned counterfactuals selected by perplexity and LLM-as-a-judge, respectively.}
\label{tab:f1_main_results}
\small
\setlength{\tabcolsep}{4pt}
\begin{tabular}{llrrrrrr}
\hline
\textbf{Condition} & \textbf{Prompt} 
& \textbf{GPT Gold} & \textbf{GPT Amb.} 
& \textbf{DeepSeek Gold} & \textbf{DeepSeek Amb.}
& \textbf{Llama Gold} & \textbf{Llama Amb.} \\
\hline
Original & Zero-shot & 0.88 & 0.44 & 0.76 & 0.39 & 0.76 & 0.39 \\
Original & Few-shot  & 0.89 & 0.47 & 0.77 & 0.41 & 0.72 & 0.38 \\
\hline
Synthetic-PPL & Zero-shot & 0.65 & 0.55 & 0.27 & 0.32 & 0.30 & 0.34 \\
Synthetic-PPL & Few-shot  & 0.66 & 0.55 & 0.28 & 0.30 & 0.28 & 0.30 \\
\hline
Synthetic-Judge & Zero-shot & 0.64 & 0.55 & 0.29 & 0.31 & 0.31 & 0.30 \\
Synthetic-Judge & Few-shot  & 0.65 & 0.54 & 0.31 & 0.31 & 0.29 & 0.28 \\
\hline
\end{tabular}
\end{table*}

The Original-Gold results show that the models can perform competitively when human annotators fully agree on the sentiment label. GPT-4.1-mini obtains the highest scores, with $0.88$ in zero-shot and $0.89$ in few-shot. DeepSeek-r1:8b reaches $0.76$ and $0.77$, while Llama3.1:8b obtains $0.76$ and $0.72$. These results indicate that, under clear human consensus, all models capture a substantial part of the affective signal, although the proprietary model remains stronger.

The Ambiguous condition produces a systematic reduction in performance. In the original dataset, GPT-4.1-mini drops to $0.44$ in zero-shot and $0.47$ in few-shot. DeepSeek-r1:8b reaches $0.39$ and $0.41$, while Llama3.1:8b reaches $0.39$ and $0.38$. This pattern supports the methodological decision to treat annotator disagreement as a diagnostic condition rather than as annotation noise. When human annotators disagree, LLMs also become less reliable, suggesting that political sentiment in social media may depend on pragmatic, contextual, or culturally situated cues that are not captured by surface polarity alone.

\subsection{Comparison with Conventional Classification}
\label{sec:results:reference}

\begin{figure}[h!]
    \centering
    \includegraphics[width=\linewidth]{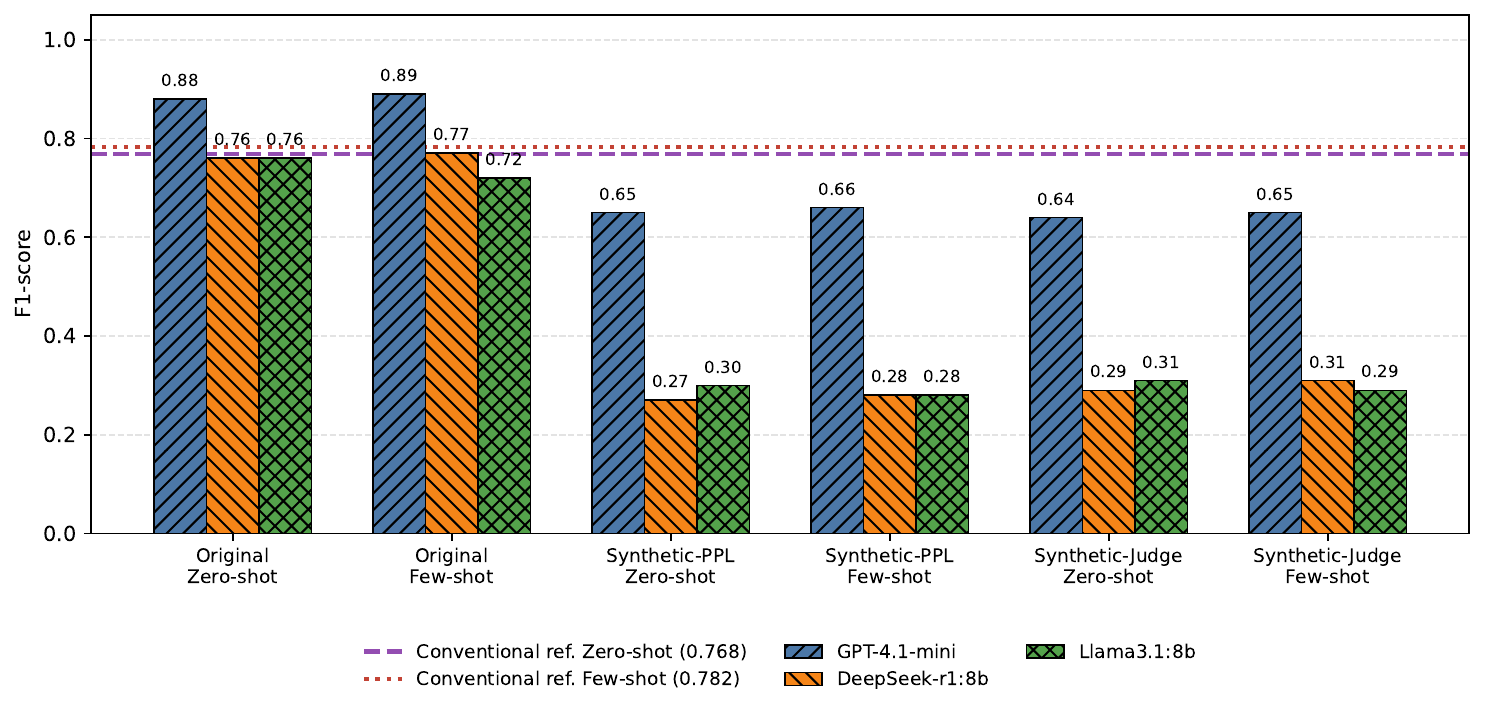}
    \caption{F1-scores on Gold instances, with contextual reference lines from conventional original-corpus sentiment classification.}
    \label{fig:f1_gold_reference}
\end{figure}

\begin{figure}[h!]
    \centering
    \includegraphics[width=\linewidth]{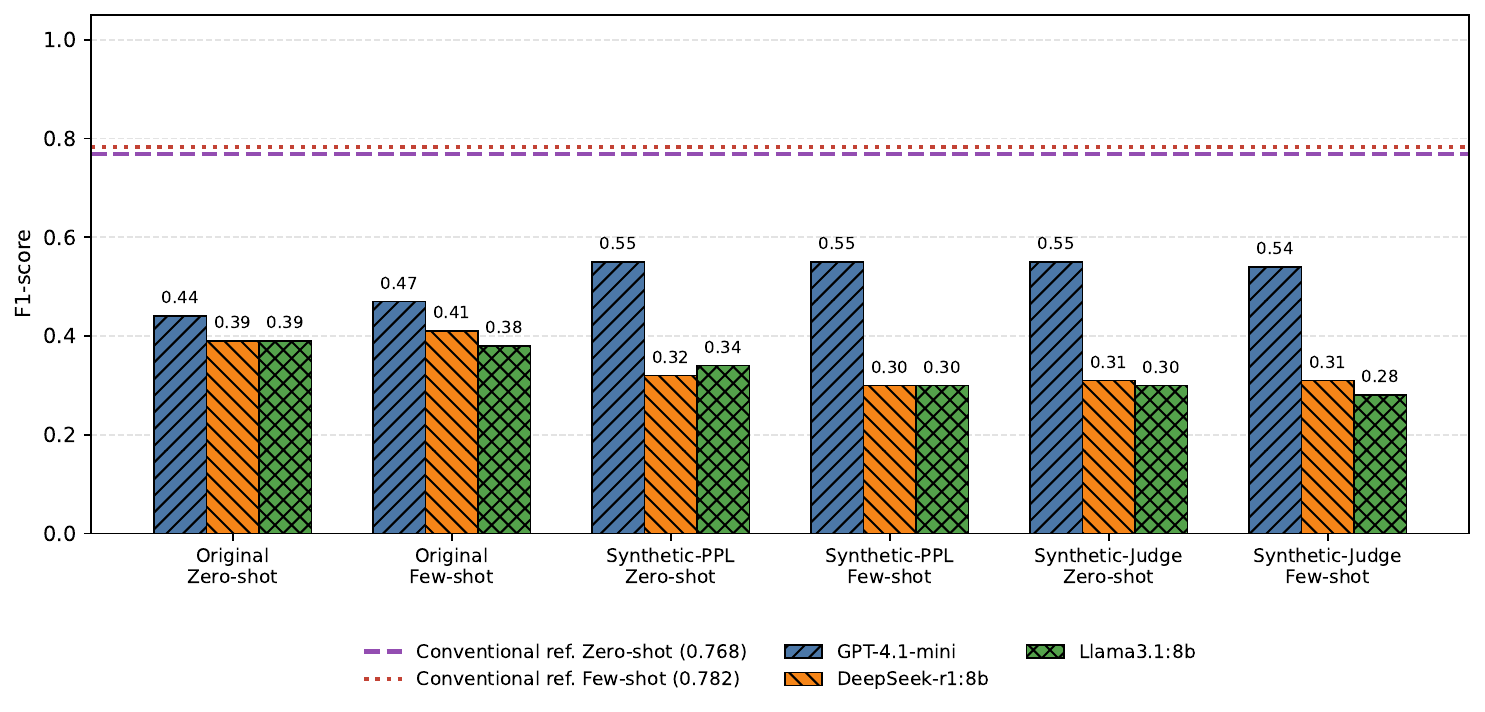}
    \caption{F1-scores on Ambiguous instances, with contextual reference lines from conventional original-corpus sentiment classification.}
    \label{fig:f1_ambiguous_reference}
\end{figure}

Figures~\ref{fig:f1_gold_reference} and~\ref{fig:f1_ambiguous_reference} provide a visual summary of the F1-scores reported in Table~\ref{tab:f1_main_results}, complemented by two horizontal reference lines. These lines correspond to a \emph{comparable conventional original-corpus sentiment classification reference} derived from prior work on the same Spanish political discourse corpus~\cite{Valverde:LLMs_sentiment:classif:26}. The reference is contextual rather than a direct baseline, since the prior evaluation was organized by platform and did not include agreement-aware or counterfactual conditions.

The reference values were computed from the mean Weighted-F1 across platforms for the two most comparable models reported in the prior benchmark, namely GPT-4o-mini and Llama-3.1. For each prompting setting, the platform-level Weighted-F1 values were first averaged within each model and then averaged across the two models. Formally, for a prompting setting $p$, the reference is computed as:

\begin{equation}
\mathrm{Ref}_{p} =
\frac{1}{2}
\left(
\frac{1}{4}\sum_{s \in S} F1_{\mathrm{GPT4o},p,s}
+
\frac{1}{4}\sum_{s \in S} F1_{\mathrm{Llama3.1},p,s}
\right),
\end{equation}

where $S=\{\mathrm{Facebook}, \mathrm{Instagram}, \mathrm{X}, \mathrm{YouTube}\}$. This gives a zero-shot reference of $0.768$ and a few-shot reference of $0.782$.

Figure~\ref{fig:f1_gold_reference} shows that Original-Gold performance for GPT-4.1-mini exceeds the conventional reference, while DeepSeek-r1:8b and Llama3.1:8b are close to or below that reference depending on the prompting setting. However, all Synthetic-Gold scores fall below the conventional reference. This is particularly evident for DeepSeek-r1:8b and Llama3.1:8b, whose Synthetic scores remain near $0.27$--$0.31$. GPT-4.1-mini also decreases under counterfactual shift, but remains substantially above the open-weight models evaluated here, with Synthetic-Gold scores between $0.64$ and $0.66$.

Figure~\ref{fig:f1_ambiguous_reference} shows a stronger contrast. All Ambiguous results remain below the conventional reference, including the original classification condition. This indicates that a conventional original-corpus benchmark can substantially overestimate the robustness of sentiment classification when it does not distinguish between high-consensus and low-consensus instances. Interestingly, GPT-4.1-mini performs better on Synthetic-Ambiguous instances than on Original-Ambiguous instances, reaching approximately $0.54$--$0.55$ under both selection strategies. This suggests that selected counterfactuals may sometimes express the intended polarity more explicitly than original posts that humans themselves judged inconsistently. The open-weight models do not show the same recovery and remain close to $0.28$--$0.34$ in Synthetic-Ambiguous settings.

\subsection{Robustness Drop Relative to Original-Gold}
\label{sec:results:drop}

To quantify robustness degradation, we compute the F1-score difference between each condition and the corresponding Original-Gold result for the same model and prompting strategy. The drop is defined as:

\begin{equation}
\Delta F1_{m,p,c} = F1_{m,p,c} - F1_{m,p,\mathrm{Original\text{-}Gold}},
\end{equation}

where $m$ denotes the model, $p$ the prompting strategy, and $c$ the evaluated condition. Under this definition, the Original-Gold condition has $\Delta F1=0$ by construction. Negative values indicate performance degradation relative to the clearest human-agreement condition.

\begin{figure*}[t]
    \centering
    \includegraphics[width=\textwidth]{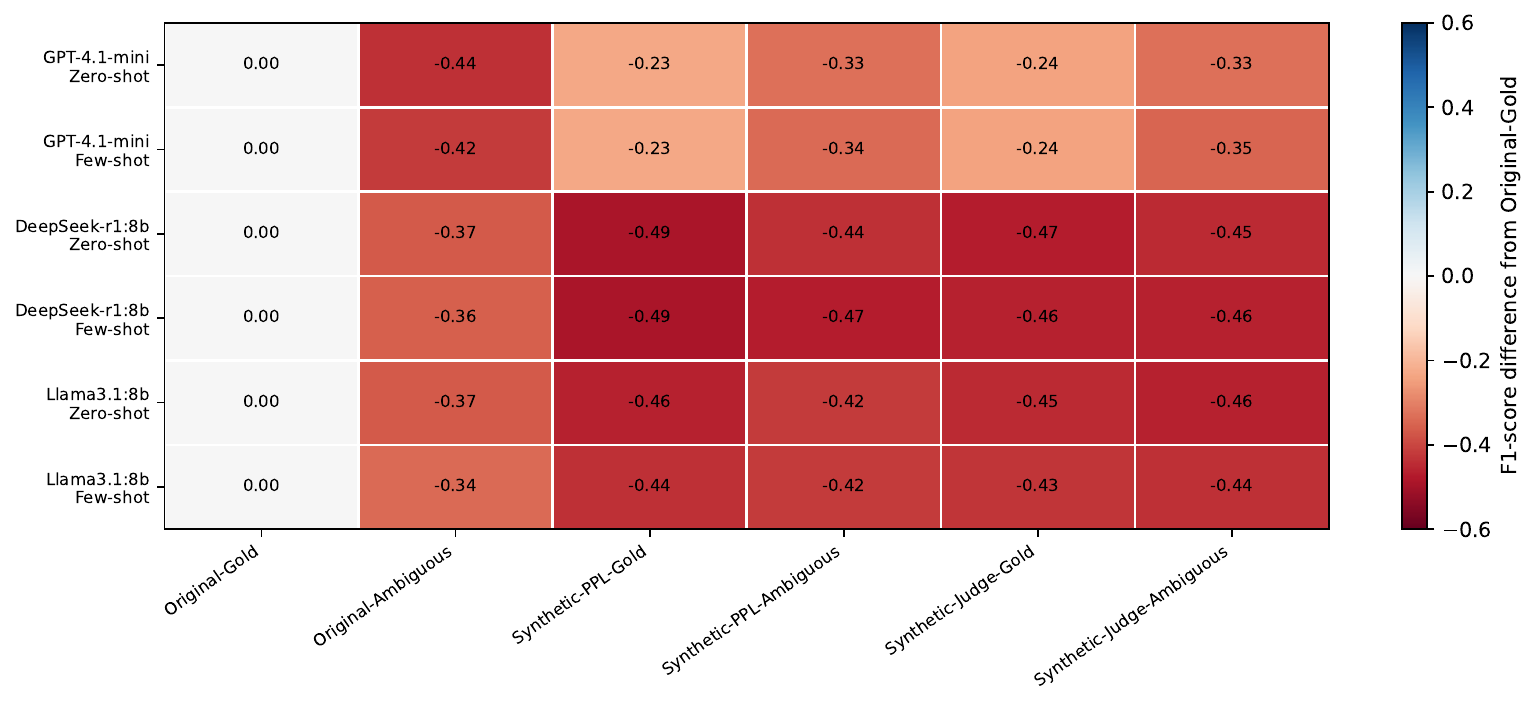}
    \caption{F1-score drop relative to Original-Gold. Negative values indicate degradation with respect to the same model and prompting strategy under full annotator agreement.}
    \label{fig:f1_drop_heatmap}
\end{figure*}

Figure~\ref{fig:f1_drop_heatmap} confirms that both ambiguity and counterfactual shift produce substantial degradation, but with different profiles across models. For Original-Ambiguous, the drop ranges from $-0.34$ to $-0.44$. GPT-4.1-mini loses $0.44$ F1 points in zero-shot and $0.42$ in few-shot. DeepSeek-r1:8b loses $0.37$ and $0.36$, while Llama3.1:8b loses $0.37$ and $0.34$. These values confirm that human disagreement is a strong indicator of model difficulty across all model families.

The Synthetic conditions reveal a sharper separation between GPT-4.1-mini and the open-weight models. For GPT-4.1-mini, Synthetic-Gold produces drops of approximately $-0.23$ to $-0.24$, regardless of whether counterfactuals are selected by perplexity or by LLM-as-a-judge. For DeepSeek-r1:8b, the Synthetic-Gold drop reaches $-0.49$ under perplexity selection and $-0.47$ or $-0.46$ under judge-based selection. Llama3.1:8b shows a similar degradation, with drops between $-0.43$ and $-0.46$. These results indicate that counterfactual perturbation exposes robustness gaps that are much less visible in Original-Gold evaluation.

The comparison between Synthetic-PPL and Synthetic-Judge also shows that the general pattern is stable across selection strategies. GPT-4.1-mini consistently remains the most robust model, while DeepSeek-r1:8b and Llama3.1:8b remain substantially more vulnerable to synthetic sentiment shifts. This suggests that the main factor driving the observed degradation is not only the specific counterfactual selection criterion, but the model's ability to recognize and preserve affective inversion under controlled rewriting.

\subsection{Effect of Few-shot Prompting}
\label{sec:results:fewshot}

We also examine whether few-shot prompting mitigates the degradation observed under ambiguity and counterfactual shift. As shown in Table~\ref{tab:f1_main_results}, the gains from few-shot prompting are small and inconsistent across models and conditions. In the Original-Gold condition, GPT-4.1-mini increases only from $0.88$ to $0.89$, and DeepSeek-r1:8b from $0.76$ to $0.77$, whereas Llama3.1:8b decreases from $0.76$ to $0.72$. A similar pattern appears in Original-Ambiguous, where GPT-4.1-mini improves from $0.44$ to $0.47$ and DeepSeek-r1:8b from $0.39$ to $0.41$, while Llama3.1:8b decreases slightly from $0.39$ to $0.38$.

The Synthetic conditions reinforce this interpretation. For GPT-4.1-mini, few-shot prompting produces almost no change under either selection strategy, moving from $0.65$ to $0.66$ in Synthetic-PPL-Gold and from $0.64$ to $0.65$ in Synthetic-Judge-Gold, while remaining effectively stable in the corresponding Ambiguous-derived subsets. DeepSeek-r1:8b shows small gains only in some cases, such as Synthetic-Judge-Gold, where it increases from $0.29$ to $0.31$, but it decreases under Synthetic-PPL-Ambiguous, from $0.32$ to $0.30$. Llama3.1:8b does not benefit consistently from few-shot examples, decreasing in most Synthetic conditions reported in Table~\ref{tab:f1_main_results}.

These results indicate that the main difficulty is not simply the absence of task examples or output-format guidance. If the errors were primarily due to prompt underspecification, few-shot prompting would be expected to produce more systematic improvements. Instead, the limited gains suggest that the performance loss is driven by the intrinsic difficulty of interpreting politically situated affective meaning under annotator ambiguity and controlled sentiment perturbation. This is also consistent with Figure~\ref{fig:f1_drop_heatmap}, where few-shot rows exhibit degradation patterns comparable to their zero-shot counterparts rather than a clear recovery toward the Original-Gold condition.

\subsection{Limitations}
\label{sec:results:limitations}

Our results should be interpreted within the scope of the proposed diagnostic design. First, the Synthetic condition was constructed through a model-assisted and model-conditioned pipeline. Counterfactual generation and judge-based selection may therefore introduce self-selection bias, since a model may select rewrites that are more compatible with its own generation style or easier for itself to classify. Consequently, the Synthetic-Judge results in Table~\ref{tab:f1_main_results} should be read as end-to-end diagnostic evidence of model behavior, rather than as externally validated evidence of counterfactual quality.

Second, the comparison between Synthetic-PPL and Synthetic-Judge partially mitigates this concern. The similar degradation patterns in Figure~\ref{fig:f1_drop_heatmap} suggest that the main robustness trends are not entirely dependent on a single selection criterion. Nevertheless, neither perplexity-based selection nor LLM-as-a-judge selection replaces human validation. Accordingly, this study does not make claims about counterfactual quality in terms of fluency, naturalness, meaning preservation, or clarity of sentiment inversion, since these dimensions were not systematically assessed.

Third, the scope of the evaluated models and experimental settings is limited. The study compares three relatively small LLMs under zero-shot and few-shot prompting, without larger frontier models, fine-tuned supervised baselines, or sentiment-specific models trained on the target domain. Thus, the observed degradation patterns should not be interpreted as definitive evidence about all LLMs or about open-weight models in general. Differences among GPT-4.1-mini, DeepSeek-r1:8b, and Llama3.1:8b may reflect interacting factors such as model scale, instruction tuning, multilingual coverage, training data composition, and alignment strategy.

Finally, the analysis does not model platform-specific behavior beyond descriptive corpus statistics. Although Table~\ref{tab:data_distribution} shows substantial variation across Facebook, Instagram, X, and YouTube, the evaluation treats agreement condition and counterfactual perturbation as the primary diagnostic axes. Platform-aware modeling is therefore left for future work. Similarly, model explanations were retained for traceability, but explanation quality was not evaluated as a main empirical axis.

\section{Conclusion}
\label{sec:conclusion}

This paper evaluated LLM robustness for Spanish political sentiment analysis through a diagnostic protocol that goes beyond conventional original-corpus classification. Rather than treating sentiment prediction as a single aggregate task, we reorganized evaluation around three conditions: Gold instances with full human agreement, Ambiguous instances with annotator disagreement, and Synthetic instances generated through sentiment-flipped counterfactual perturbation. The results show that conventional sentiment classification can provide an incomplete picture of model reliability. In the  Gold subset, the models achieved comparatively strong F1-scores, with GPT-4.1-mini exceeding the contextual conventional reference derived from prior platform-based evaluation. However, performance decreased substantially when the same task was evaluated under human ambiguity and counterfactual shift. This indicates that high performance on standard sentiment benchmarks does not necessarily imply robustness when affective meaning is culturally situated, pragmatically indirect, or unstable even among human annotators.

The main empirical finding is that annotator ambiguity and counterfactual perturbation expose different forms of model fragility. Ambiguous instances produced systematic drops across all models, suggesting that human disagreement is a meaningful signal of interpretive difficulty rather than merely annotation noise. Synthetic instances further revealed robustness gaps that were especially pronounced for the open-weight models. GPT-4.1-mini remained the most stable model across conditions, although it also degraded outside the Gold setting. DeepSeek-r1:8b and Llama3.1:8b performed reasonably on clear original instances but showed stronger degradation under counterfactual shift. These results suggest that politically situated sentiment analysis in Spanish requires evaluation protocols that explicitly account for disagreement, perturbation, and contextual affective cues.

The comparison with conventional sentiment classification is central to the contribution of this work. Standard evaluations organized by platform or aggregate performance remain valuable for benchmarking models under ordinary classification conditions. However, they do not fully address whether models preserve sentiment interpretation when language becomes ambiguous or when sentiment is intentionally inverted while preserving topic and political context. Our results therefore do not replace conventional sentiment analysis; rather, they complement it by showing where conventional evaluation may overestimate robustness. This distinction is particularly important for political discourse, where sentiment is often expressed through irony, emojis, hashtags, local references, implicit stance, and platform-specific communicative conventions.

This study has several limitations. First, the Synthetic condition was constructed through a model-assisted counterfactual pipeline, and therefore should be interpreted as a diagnostic stress condition rather than as a fully independent human-authored benchmark. Second, although counterfactual generation and selection were supported by perplexity and LLM-as-a-judge criteria, the present paper does not provide a full human validation of counterfactual quality. Third, we retained model explanations as part of the structured output, but we did not include explanation quality as a main empirical axis in this version of the paper. A more rigorous analysis of explanation faithfulness, contextual adequacy, and reasoning consistency remains necessary before making stronger claims about explanatory robustness. Fourth, the analysis did not model platform-specific behavior beyond reporting corpus composition, even though the distribution of sentiment and ambiguity differs across Facebook, Instagram, X, and YouTube. Finally, the study is limited to three LLMs and does not include a broader family of multilingual, sentiment-specific, or instruction-tuned models.

Future work should address these limitations in four directions. First, the counterfactual subset should be strengthened through human validation, including judgments of fluency, naturalness, meaning preservation, and clarity of sentiment inversion. Second, explanation quality should be evaluated systematically, ideally combining human assessment with constrained model-assisted rubrics. Third, the protocol should be extended to platform-aware analysis, examining whether models behave differently across social media environments with distinct linguistic, demographic, and interactional patterns. Fourth, future experiments should include additional affective dimensions beyond polarity, such as emotion, sarcasm, aggressiveness, polarization, uncertainty, and stance. These extensions would allow a more complete characterization of political affect in Spanish social media discourse.

Overall, this work contributes to the study of natural language reasoning in Spanish by showing that sentiment analysis becomes substantially more demanding when evaluated through human ambiguity and controlled counterfactual shift. The proposed protocol provides a practical way to examine whether LLMs merely assign labels or whether they remain stable when affective meaning depends on culturally and pragmatically situated cues. In doing so, it also contributes to a better understanding of Spanish-language social platforms as complex spaces of political expression, where sentiment cannot always be reduced to surface polarity. The dataset annotations, evaluation artifacts, and supporting materials are publicly available in the repository: \url{https://github.com/jvalverr/sentimex-llm-benchmark/tree/main/cultural_reasoning}.

\bibliographystyle{IEEEtran}
\bibliography{references}

\end{document}